\documentclass[10pt,twocolumn,letterpaper]{article}

\usepackage{multirow}
\usepackage{tabularx}
\usepackage{float}
\usepackage{stfloats}
\usepackage[section]{placeins}
\usepackage[pagenumbers]{cvpr} % To force page numbers, e.g. for an arXiv version

\usepackage{microtype}
\usepackage{graphicx}
\usepackage{subcaption}
\usepackage{booktabs} % for professional tables

\usepackage{xcolor}
\usepackage{color}
\usepackage{multirow}
\usepackage{tabularx}
\usepackage{float}
\usepackage{algorithm}
\usepackage{algpseudocode}
\usepackage{setspace}
\usepackage{bm}
\usepackage[pagebackref=true,breaklinks=true,letterpaper=true,colorlinks,bookmarks=false]{hyperref}

\definecolor{cvprblue}{rgb}{0.21,0.49,0.74}
\definecolor{cvprblue}{rgb}{0.21,0.49,0.74}
\definecolor{lightpurple}{rgb}{0.902, 0.863, 0.922}
\definecolor{lightgreen}{rgb}{0.863, 0.902, 0.863}
\definecolor{lightpink}{rgb}{0.980, 0.941, 0.941}

\newcommand{\best}[1]{\textbf{\textcolor{red}{#1}}}
\newcommand{\sbest}[1]{\underline{\textcolor{blue}{#1}}}
\def\confName{CVPR}
\def\confYear{2026}

\title{Noise-Started One-Step Real-World Super-Resolution via LR-Conditioned SplitMeanFlow and GAN Refinement}

\author{
Wei Zhu$^{1}$ \qquad
Kai Zhang$^{2,}\thanks{Corresponding authors.}$ \qquad
Yu Zheng$^{1}$ \qquad
Lei Luo$^{1,*}$ \qquad
Yong Guo$^{3}$ \qquad
Jian Yang$^{1,2,*}$ \\
$^{1}$Nanjing University of Science and Technology \qquad
$^{2}$Nanjing University  \\
$^{3}$South China University of Technology\\
\url{https://github.com/wzhu121/SMFSR}
}

\begin{document}

\setlength{\abovedisplayskip}{3pt}
\setlength{\belowdisplayskip}{2pt}
\maketitle

\begin{abstract}
Pre-trained text-to-image (T2I) diffusion models have shown strong potential for real-world image super-resolution (Real-ISR), owing to their noise-started generation process that enables realistic texture synthesis and captures the one-to-many nature of super-resolution. However, diffusion-based Real-ISR methods still face a fundamental efficiency-quality trade-off. Multi-step methods generate high-quality results by iteratively denoising random Gaussian noise under LR conditioning, but suffer from slow sampling. Recent one-step methods greatly improve efficiency, yet they typically replace noise-started generation with direct LR-to-HR restoration, which weakens stochasticity and limits realistic detail synthesis.
To address this issue, we propose \textbf{SMFSR}, a noise-started one-step Real-ISR framework via \textbf{LR-conditioned SplitMeanFlow} and \textbf{GAN refinement}. SMFSR preserves the random-noise starting point of diffusion models and learns a direct noise-to-HR mapping conditioned on the LR image. To this end, Interval Splitting Consistency distills the multi-step generative trajectory into a single average-velocity prediction, enabling efficient one-step generation. To compensate for the reduced opportunity for progressive refinement, we further introduce a GAN refinement stage, where a DINOv3-based discriminator enhances realistic texture synthesis and variational score distillation aligns the generated outputs with the natural image distribution under a frozen diffusion teacher. Extensive experiments demonstrate that SMFSR achieves state-of-the-art perceptual quality among one-step diffusion-based Real-ISR methods while retaining fast single-step inference.
\end{abstract}    
\section{Introduction}

\begin{figure}[t!]
    \centering
    \begin{subfigure}[b]{0.495\textwidth}
        \centering
        \includegraphics[width=\textwidth]{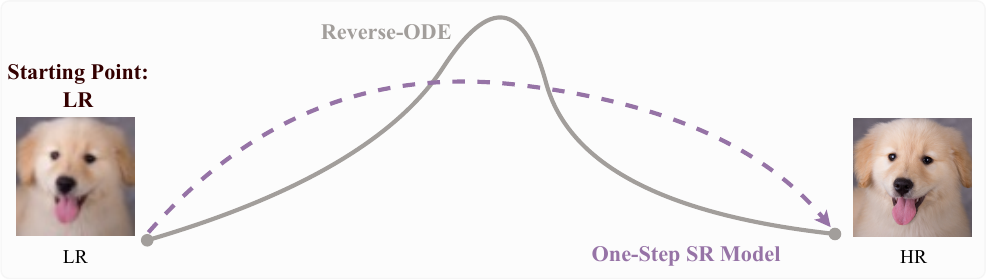}
        \caption{\textbf{Vanilla One-Step Diffusion for Real-ISR.} The HR image is directly restored from the LR input in a single step, improving efficiency but sacrificing stochasticity and perceptual quality.}
    \end{subfigure}
    \begin{subfigure}[b]{0.495\textwidth}
        \centering
        \includegraphics[width=\textwidth]{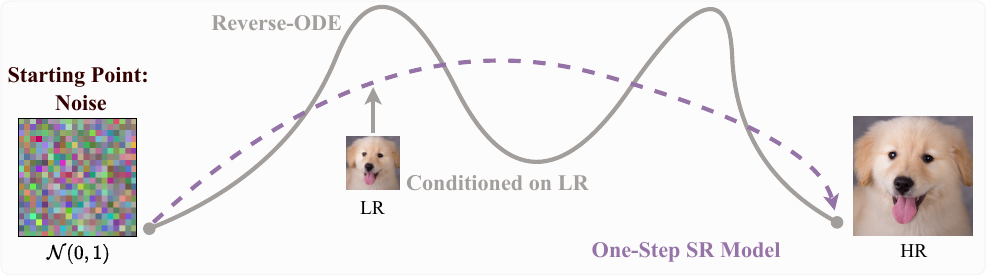}
        \caption{\textbf{Our Noise-Started One-Step Diffusion for Real-ISR.} Starting from random Gaussian noise, our method performs one-step HR generation with LR conditioning, achieving efficient inference and strong perceptual quality.}
    \end{subfigure}
    \caption{Unlike many diffusion-based one-step methods that directly map LR inputs to HR outputs, our method preserves the noise-started generation paradigm and generates HR images from random noise in a single step under LR conditioning. \textit{This design retains the stochastic generative capacity of diffusion models while enabling fast inference.}}
    \vspace{-3mm}
    \label{pic:teaser}
\end{figure}

\label{sec:intr}

Image super-resolution (ISR) aims to recover a high-resolution (HR) image from its low-resolution (LR) observation. Classical SR methods~\cite{2017Enhanced,2021Approaching} usually assume simple and predefined degradation models, which limits their generalization to practical scenarios. Real-world image super-resolution (Real-ISR)~\cite{2021Designing,2021Real} instead addresses complex and unknown degradations, and has therefore become a more realistic and challenging setting. Recently, pre-trained text-to-image (T2I) diffusion models have shown remarkable potential for Real-ISR~\cite{yang2023pasd,2024stablesr,wu2024seesr,supir}. By generating images from random noise under LR conditioning, these models provide strong generative priors for synthesizing realistic textures and modeling the one-to-many nature of SR.

Existing diffusion-based Real-ISR methods still face a fundamental efficiency-quality trade-off. Multi-step methods~\cite{supir,wu2024seesr} preserve the original diffusion generation process: they start from random Gaussian noise and iteratively denoise it under LR guidance to obtain the HR output. This noise-started iterative process can produce photo-realistic images with rich details, but it requires many sampling steps and thus suffers from slow inference. To improve efficiency, recent one-step methods~\cite{2024OSEDiff,2024sinsr} distill diffusion priors into single-step restoration networks. However, most of them initiate restoration directly from the LR input, as illustrated in Fig.~\ref{pic:teaser}(a). Although efficient, this design changes the diffusion paradigm from noise-to-image generation into direct LR-to-HR restoration, which weakens stochasticity and limits the ability to synthesize diverse and realistic high-frequency details.

This raises a natural question: \emph{Can one-step Real-ISR retain the noise-started generation paradigm of diffusion models while achieving fast inference?} Answering this question requires learning a direct mapping from random Gaussian noise to HR images under LR conditioning. However, this is non-trivial, since the original diffusion trajectory relies on progressive denoising over multiple steps. Compressing such a long generative trajectory into a single prediction reduces the opportunity for gradual structure formation and detail refinement. Consequently, a one-step model may recover the overall content but struggle to generate realistic textures and fine perceptual details. An effective solution should therefore satisfy two requirements: preserving the noise-to-HR formulation for stochastic generation, and introducing additional perceptual supervision to compensate for the loss of progressive refinement.

To this end, we propose \textbf{SplitMeanFlow for Super-Resolution (SMFSR)}, a noise-started one-step Real-ISR framework based on \textbf{LR-conditioned SplitMeanFlow} and \textbf{GAN refinement}, as shown in Fig.~\ref{pic:teaser}(b). SMFSR starts from random Gaussian noise and generates HR images in a single step conditioned on the LR input. Its core component is SplitMeanFlow, which predicts an average velocity field over a time interval instead of estimating a conventional instantaneous denoising direction. With Interval Splitting Consistency (ISC), SMFSR distills the multi-step generative trajectory into a single average-velocity prediction from noise to the HR latent space. This formulation preserves the random-noise starting point of diffusion models while enabling efficient one-step inference.

Despite its efficiency, ISC-based one-step generation may still lack sufficient detail refinement compared with multi-step diffusion sampling. This limitation is expected, because a single average-velocity prediction has to approximate a long generative trajectory and thus provides fewer opportunities for progressive texture formation. To address this issue, we further introduce a GAN-based refinement stage. Specifically, we use a DINOv3-based discriminator~\cite{dinov3} to enhance structural realism and texture synthesis by leveraging its strong self-supervised visual representations. We further adopt variational score distillation (VSD)~\cite{vsd}, which aligns the one-step outputs with the natural image distribution using a frozen pre-trained diffusion teacher and a trainable regularizer. A reconstruction loss is also adopted to maintain fidelity to the LR observation. In this way, SMFSR forms a noise-to-structure-to-detail generation pipeline: LR-conditioned SplitMeanFlow first enables noise-started one-step HR generation, and GAN refinement further enhances perceptual realism and fine details.

Extensive experiments on both synthetic and real-world benchmarks demonstrate that SMFSR achieves state-of-the-art perceptual quality among one-step diffusion-based Real-ISR methods while requiring only a single inference step.

Our contributions are summarized as follows:
\begin{itemize}
  \item We propose a noise-started one-step Real-ISR framework that preserves the random-noise generation paradigm of diffusion models, in contrast to existing one-step methods that directly restore HR images from LR inputs.

  \item We introduce LR-conditioned SplitMeanFlow for Real-ISR and use Interval Splitting Consistency to learn a direct one-step average-velocity mapping from random Gaussian noise to HR images.

  \item Extensive experiments on synthetic and real-world benchmarks show that SMFSR achieves superior perceptual quality over existing one-step diffusion-based Real-ISR methods while maintaining efficient single-step inference.
\end{itemize}
\section{Related Work}
\label{sec:related_work}

\noindent
\textbf{Diffusion and Flow-based Generative Models.}
Diffusion models~\cite{diffusion_beat_gan} have achieved remarkable success in image generation by progressively denoising random Gaussian noise into realistic images. Despite their strong generative capacity, standard diffusion models usually require many sequential sampling steps, leading to high inference cost. To improve sampling efficiency, flow matching~\cite{flow_match} formulates generative modeling as learning a time-dependent velocity field that transports samples from a simple prior distribution to the data distribution through ordinary differential equations (ODEs). This deterministic transport formulation provides an effective alternative to iterative stochastic denoising and has been widely adopted in recent generative frameworks, including Rectified Flow~\cite{rec-flow}, SD3~\cite{sd3}, and Flux~\cite{flux2024}. These methods demonstrate improved sampling efficiency and controllability in text-to-image generation.
Recent image restoration and super-resolution methods have also begun to exploit flow-based generative priors. For example, DiT4SR~\cite{dit4sr} adopts a diffusion transformer architecture for real-world SR, while TSD-SR~\cite{dong2024tsd} introduces target score distillation to improve one-step restoration quality. However, most existing efficient Real-ISR methods still formulate one-step inference as direct restoration from LR inputs, rather than preserving the random-noise starting point of generative models. 

\noindent
\textbf{One-Step Diffusion-based Real-ISR.}
Pre-trained text-to-image diffusion models provide powerful generative priors for real-world image super-resolution. Multi-step diffusion-based Real-ISR methods~\cite{supir,wu2024seesr,yang2023pasd,2024stablesr} typically start from random Gaussian noise and iteratively denoise it under LR guidance. This noise-started iterative generation process can synthesize realistic textures and model the one-to-many nature of SR, but it also incurs substantial sampling cost. To reduce inference time, recent methods attempt to distill diffusion priors into one-step models. OSEDiff~\cite{2024OSEDiff} starts restoration from the LR image and introduces variational score distillation~\cite{vsd} to transfer the generative prior of a pre-trained diffusion model. TSD-SR~\cite{dong2024tsd} employs target score distillation to provide more reliable training signals for perceptual restoration. CTMSR~\cite{CTMSR} further uses consistency training to learn a deterministic one-step mapping from degraded LR inputs to HR outputs. These methods significantly improve efficiency, but they largely convert diffusion-based SR from noise-to-image generation into direct LR-to-HR restoration.
Such a paradigm shift limits the stochastic generative capacity inherited from diffusion models. Since the HR output is directly determined by the LR input, existing one-step methods have limited ability to produce diverse plausible details for the same LR observation and often remain perceptually inferior to multi-step diffusion methods~\cite{chen2025codsr,kong2025nsarm}. 
\section{Method}
\label{sec:method}
% ----------------------- figure method ----------------------
\begin{figure*}[t]
    \centering
    \includegraphics[width=1.0\linewidth]{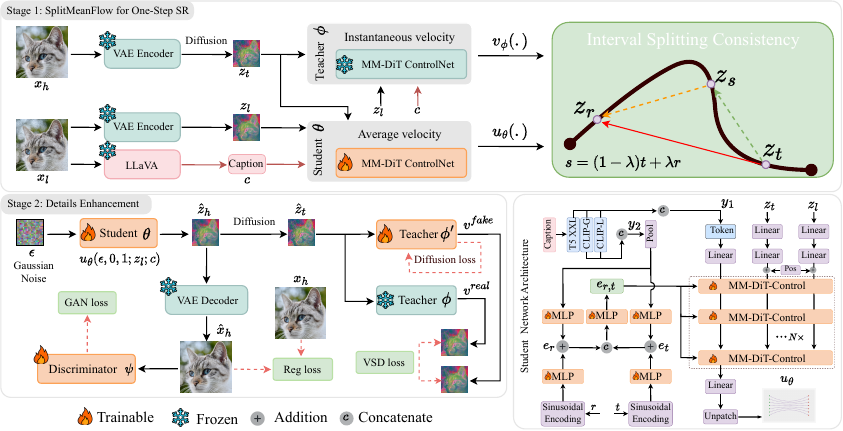}
    % \vspace{2mm}
    \caption{\textbf{Training framework of SMFSR.}
SMFSR is trained in two stages: (1) \textbf{Noise-started one-step generation with LR-conditioned SplitMeanFlow}, where Interval Splitting Consistency trains a student model to predict the average velocity from random noise to the HR latent, enabling one-step noise-to-HR sampling with $r=0$ and $t=1$; and (2) \textbf{GAN-based detail refinement}, where adversarial loss, VSD loss, and regularization loss are introduced to enhance perceptual details and visual realism.}
    \label{fig:method}
    \vspace{-3mm}
\end{figure*}

\subsection{Preliminaries}
\label{sec:preliminaries}
% ------------------flow match models-----------------
\paragraph{Flow Matching.}
Flow Matching~\cite{song2021scorebased, rec-flow} is a generative modeling framework that learns a time-dependent velocity field to transport samples from a simple prior distribution to the data distribution. Let $\epsilon \sim p_{\text{prior}}(\epsilon)$ denote a noise sample and $x \sim p_{\text{data}}(x)$ denote a data sample. A simple linear interpolation path connecting $\epsilon$ and $x$ can be defined as:
\begin{equation}
    z_t = (1-t) x + t \epsilon, \quad t \in [0,1],
\end{equation}
where $z_0 = x$ and $z_1 = \epsilon$, ensuring that the path starts from noise at $t=1$ and reaches the data sample at $t=0$.
The instantaneous velocity along this path is defined as:
\begin{equation}
    v_t = \frac{dz_t}{dt} = \epsilon - x.
\end{equation}
A neural network $v_\theta(z_t, t)$ is trained to predict this velocity field by minimizing the expected squared error:
\begin{equation}
    \mathcal{L}_{\text{FM}}(\theta) =\mathbb{E}_{x,\, \epsilon,\, t \sim \mathcal{U}(0,1)}\left\| v_\theta(z_t, t) - (\epsilon - x) \right\|^2.
\end{equation}
After training, new samples can be generated by integrating the corresponding ordinary differential equation (ODE):
\begin{equation}
    \frac{dz_t}{dt} = v_\theta(z_t, t),
\end{equation}
which gradually transforms noise samples into data samples.

\paragraph{MeanFlow with Interval Splitting Consistency.}
Unlike standard flow matching, which learns the instantaneous velocity field $v(z_t, t)$, MeanFlow~\cite{meanflow} models the average velocity over a time interval $[r,t]$:
\begin{equation}
u(z_t, r, t) = \frac{1}{t-r} \int_{r}^{t} v(z_\tau, \tau)\, d\tau.
\end{equation}
A key property of MeanFlow is the flow identity, which relates the average velocity to the instantaneous velocity:
\begin{equation}
u(z_t, r, t) = v(z_t, t) - (t - r)\frac{d}{dt}u(z_t, r, t).
\end{equation}
This relation enables training without direct supervision of the true instantaneous velocity.
Based on this identity, the model is trained by minimizing
\begin{equation}
\mathcal{L}(\theta) = \mathbb{E}_{t,r,z_t}  \| u_\theta(z_t, r, t) - \text{stopgrad}(u_{\text{target}}) \|^2 ,
\end{equation}
where $u_{\text{target}} = v_t - (t-r)\left( v_t \cdot \partial_z u_\theta + \partial_t u_\theta \right)$, and $\operatorname{stopgrad}(\cdot)$ denotes the stop-gradient operation.

To avoid explicit differential operators, SplitMeanFlow~\cite{splitmeanflow} reformulates the objective using an algebraic consistency constraint. Specifically, for any $r \leq s \leq t$, the additivity of integrals leads to the Interval Splitting Consistency condition:
\begin{equation}
\small{(t-r)u(z_t, r, t) = (s-r)\,u(z_s, r, s) + (t-s)\,u(z_t, s, t).}
\label{eq:split}
\end{equation}
This formulation establishes SplitMeanFlow as a direct and general framework for learning average velocity fields. Moreover, by eliminating the need for Jacobian-vector product (JVP) computations, it substantially improves computational efficiency and leads to more stable training. This property is particularly suitable for one-step Real-ISR, because the full trajectory from random noise to the HR output can be represented by a single average velocity over the interval $[0,1]$.

\subsection{LR-Conditioned SplitMeanFlow for Noise-Started Real-ISR}
\label{sec:training_objective}

%  ---------------- Student Network Architecture -----------
\noindent
\textbf{Student Network Architecture.}
Our goal is to build a noise-started one-step Real-ISR framework that preserves the generative formulation of diffusion models. Instead of directly restoring an HR image from the LR input, our model starts from random Gaussian noise and generates the HR latent in a single step under LR conditioning. To this end, we introduce SplitMeanFlow into diffusion-based SR and learn an LR-conditioned average-velocity field from noise to the HR latent, thereby retaining the stochastic generative capacity of diffusion models.

Our student model is built upon DiT4SR~\cite{dit4sr}, which injects LR information into the native Diffusion Transformer (DiT)~\cite{dit} blocks instead of relying on an external control branch. To support SplitMeanFlow, we redesign timestep conditioning to take two timesteps, $r$ and $t$, as illustrated in Fig.~\ref{fig:method}. Given the caption $c$ extracted from the LR input, we encode it with three pre-trained text encoders, including CLIP-L~\cite{2021Learning}, CLIP-G~\cite{clip_g}, and T5-XXL~\cite{t5xxl}. This produces two types of representations, denoted as $y_1$ and $y_2$. Specifically, $y_1$ is projected by a linear layer and used as the text tokens of DiT, while $y_2$ is obtained from the two CLIP encoders and pooled into a global representation. For each timestep, we add positional embeddings and combine them with $y_2$ to obtain the timestep embeddings $\mathbf{e}_r$ and $\mathbf{e}_t$. These two embeddings are then fused into the final timestep embedding $\mathbf{e}_{r,t}$, which modulates the internal features of DiT.

For image conditioning, we encode the LR image $x_l$ and the HR image $x_h$ into the latent space using a pre-trained VAE encoder, yielding $z_l$ and $z_h$, respectively. We sample random Gaussian noise $\epsilon$ and construct the noisy latent $z_t$ along the flow path between $\epsilon$ and $z_h$. The LR latent $z_l$ and the noisy latent $z_t$ are patchified and linearly projected into input tokens, with the same positional embedding added to both. Together with the text tokens $y_1$, these tokens are processed by $N$ stacked MM-DiT-Control blocks. The output tokens are then unpatchified to produce the predicted average velocity field $u_{\theta}$.

\noindent
\textbf{Interval Splitting Consistency Loss.}
To enable one-step generation from random noise to the HR latent conditioned on $z_l$ and $c$, we learn an average velocity field $u_\theta$ that describes the transition between two time points $r$ and $t$. In particular, when $r=0$ and $t=1$, the learned field covers the full trajectory and directly maps random Gaussian noise to the HR latent in a single inference step.

Based on the Interval Splitting Consistency in Eq.~\eqref{eq:split}, we formulate the LR-conditioned consistency objective as:
\begin{flalign}
&\qquad\qquad \mathcal{L}_{\text{ISC}} =
\big\|(t-r)u_\theta(z_t,r,t;z_l,c) - \label{eq:scm_raw} \\
&\left[(s-r)u_\theta(z_s,r,s;z_l,c)
+ (t-s)u_\theta(z_t,s,t;z_l,c)\right]\big\|^2,
\nonumber
\end{flalign}
where $u_\theta(z_t,r,t;z_l,c)$ denotes the student network parameterized by $\theta$, which predicts the average velocity over the interval $[r,t]$ conditioned on the LR latent $z_l$ and caption $c$. The intermediate state $z_s$ is obtained by backward integration from $z_t$:
\begin{equation}
    z_s = z_t - (t-s)u_\theta(z_t,s,t;z_l,c).
\end{equation}
For stable optimization, we define $\lambda=(t-s)/(t-r)$, such that $s=(1-\lambda)t+\lambda r$. Eq.~\eqref{eq:scm_raw} can then be rewritten as
\begin{flalign}
&\qquad\qquad \mathcal{L}_{\text{ISC}} =
\big\|u_\theta(z_t,r,t;z_l,c) - \text{stopgrad} \label{eq:scm}\\
&\left[(1-\lambda)u_\theta(z_s,r,s;z_l,c)
+ \lambda u_\theta(z_t,s,t;z_l,c)\right]\big\|^2.
\nonumber
\end{flalign}
This objective enforces interval-level trajectory consistency: the average velocity over the long interval $[r,t]$ should match the length-weighted combination of the average velocities over its two sub-intervals $[r,s]$ and $[s,t]$. By repeatedly matching long- and short-interval predictions, the student learns the full noise-to-HR trajectory under LR conditioning. During inference, setting $r=0$ and $t=1$ enables one-step noise-started HR generation.

\noindent
\textbf{Boundary Consistency Loss.}
Although ISC provides self-consistency across intervals, it does not by itself anchor the learned velocity field to a valid generative trajectory. To prevent training drift and stabilize degenerate intervals, we introduce a boundary consistency loss using a flow-matching teacher parameterized by $\phi$. When the interval collapses to a single time point, \ie, $r=t$, the average velocity predicted by the student should match the instantaneous velocity predicted by the teacher:
\begin{equation}
    u_{\theta}(z_t,t,t;z_l,c)=v_{\phi}^{w}(z_t,t;z_l,c),
\end{equation}
where $v_{\phi}^{w}$ denotes the classifier-free guidance velocity with guidance scale $w$:
\begin{equation}
    v_{\phi}^{w}(z_t,t;z_l,c)
    =
    w v_{\phi}^{\text{cond}}(z_t,t;z_l,c)
    +
    (1-w)v_{\phi}^{\text{uncond}}(z_t,t;z_l).
\end{equation}
The overall training procedure of LR-conditioned SplitMeanFlow is summarized in Algorithm~\ref{alg: overall training process}.

\subsection{GAN-based Detail Refinement}
\label{sec:detailed_refinement}

After training with Interval Splitting Consistency in the first stage, the student model can generate the HR latent from random Gaussian noise in a single step by setting $r=0$ and $t=1$:
\begin{equation}
    \hat{z}_h = \epsilon - u_{\theta}(\epsilon,0,1;z_l,c).
\end{equation}
Although this formulation enables efficient noise-started generation, it compresses a multi-step generative trajectory into a single average-velocity prediction. As analyzed in Section~\ref{sec:limited_detail_recovery_main}, such trajectory compression reduces the opportunity for progressive texture enhancement. Consequently, the student can recover the global structure effectively but may produce weaker high-frequency details than the multi-step teacher.

To compensate for this limitation, we introduce a second-stage GAN-based detail refinement strategy. In this stage, the student model is further optimized with variational score distillation, adversarial supervision, and reconstruction regularization. This design allows LR-conditioned SplitMeanFlow to provide efficient one-step noise-to-HR generation, while GAN refinement enhances perceptual realism and fine textures.

\begin{algorithm}[t]
    % \caption{Overall training procedure of CTMSR.}
    \caption{Interval Splitting Consistency in \textit{stage 1}.}
    \label{alg: overall training process}
    % \vspace{-5mm}
    \begin{algorithmic}[1]
    \setstretch{1.10}
        \Require VAE encoder $\bm{\mathcal{E}}$,
        pre-trained teacher model $\bm{v}_{\bm{\phi}}(\cdot)$, student model $\bm{u}_{\bm{\theta}}(\cdot)$ initialized by $\bm{v}_{\bm{\phi}}(\cdot)$, 
       training dataset $(X_l, X_h)$, branch probability $p$.
        % \State $\textbf{Stage 1: Consistency Training for One-Step SR}$
        \While{not converged}
            \State Sample $x_l, x_h \sim (X_l, X_h)$
            \State Sample time points $r$, $t$ such that $0 \leq r \leq t \leq 1$
            \State Sample $\lambda \sim \mathcal{U}(0,1)$, and set $s=(1-\lambda)t+\lambda r$
            \State Sample prior $\epsilon \sim \mathcal{N}(0,1)$, sample $q \sim \mathcal{U}(0,1)$
            \State $z_l \leftarrow \bm{\mathcal{E}}(x_l)$, $z_h \leftarrow \bm{\mathcal{E}}(x_h)$
            \State Compute the point at time $t$: $z_t=(1-t)z_h+t\epsilon$
            % \If{$u < p$} \Comment{Compute boundary condition:}
            \If{$q < p$} \Comment{\textcolor{blue!50}{Splitting consistency}}
                \State $u_2 = \bm{u_\theta} (z_t, s, t; z_l, c)$
                \State $z_s = z_t - (t - s) u_2$
                \State $u_1 = \bm{u_\theta} (z_s, r, s; z_l, c)$
                \State $u_{\text{target}} = (1 - \lambda)u_1 + \lambda u_2$
                \State $\mathcal{L}_{\text{ISC}}
                = \left\lVert \mathbf{u}_{\theta}(z_t, r, t; z_l, c)
                - \operatorname{stopgrad}\!\left(u_{\text{target}}\right) \right\rVert^2$

            \Else \Comment{\textcolor{blue!50}{Boundary consistency}}
                \State $\mathcal{L}_\text{ISC} = \left\lVert\bm{u_\theta}(z_t, t, t; z_l, c) -  \bm{v_\phi}(z_t,t;z_l,c)\right\rVert^2$
            \EndIf
            \State Update $\theta$ using gradient descent step on $\nabla_\theta \mathcal{L}_{\text{ISC}}$
        
        \EndWhile
        \State \textbf{return} The student model $\bm{u}_{\bm{\theta}}(\cdot)$.
    \end{algorithmic}
\end{algorithm}

\noindent
\textbf{Variational Score Distillation.}
Variational Score Distillation (VSD)~\cite{vsd} aligns the distribution of generated images with the natural image distribution by optimizing a KL-divergence objective. We use a frozen pre-trained diffusion teacher parameterized by $\phi$ to provide the target score, and introduce a trainable regularizer parameterized by $\phi'$ to adaptively guide the student. The gradient with respect to the student parameters $\theta$ is formulated as:
\begin{equation}
\nabla_\theta \mathcal{L}_{\text{VSD}}
=
\mathbb{E}_{t,\epsilon}
\left[
\omega(t)
\left(
v_{\phi}(\hat z_t,t;z_l,c)
-
v_{\phi'}(\hat z_t,t;z_l,c)
\right)
\frac{\partial \hat{z}_t}{\partial \theta}
\right],
\label{eq:vsd_g}
\end{equation}
where $\hat{z}_t=(1-t)\hat{z}_h+t\epsilon$ is the noisy latent, $\epsilon\sim\mathcal{N}(0,\mathbf{I})$ denotes Gaussian noise, and $\omega(t)$ is a time-dependent weighting function. The trainable regularizer $\phi'$ is initialized from $\phi$ and updated with the standard diffusion objective:
\begin{equation}
    \mathcal{L}_{\text{diff}}
    =
    \mathbb{E}_{t,\epsilon}
    \left\|
    v_{\phi'}(\hat{z}_t,t;z_l,c)
    -
    (\epsilon-\hat{z}_h)
    \right\|^2.
\end{equation}

\noindent
\textbf{GAN Loss.}
To further enhance texture realism and structural coherence, we introduce adversarial supervision. Considering the strong visual representation and discriminative ability of DINOv3~\cite{dinov3}, we adopt a DINOv3-based discriminator to provide stable adversarial training signals. This discriminator encourages the student to synthesize perceptual details that are difficult to fully recover through a single average-velocity prediction. The adversarial objectives are defined as:
\begin{align}
\mathcal{L}_{\text{adv}}^{\mathcal{G}}
&= -\mathbb{E}_{\hat{x}_h}
\left[
\mathcal{D}_{\psi}(\hat{x}_h)
\right],
\label{eq:adv_g} \\
\mathcal{L}_{\text{adv}}^{\mathcal{D}}
&=
\mathbb{E}_{x_h}
\left[
\mathrm{max}(0,1-\mathcal{D}_{\psi}(x_h))
\right]
\nonumber \\
&\quad+
\mathbb{E}_{\hat{x}_h}
\left[
\mathrm{max}(0,1+\mathcal{D}_{\psi}(\hat{x}_h))
\right],
\label{eq:adv_d}
\end{align}
where $\mathcal{L}_{\text{adv}}^{\mathcal{G}}$ and $\mathcal{L}_{\text{adv}}^{\mathcal{D}}$ are used to optimize the student parameters $\theta$ and discriminator parameters $\psi$, respectively. The image $\hat{x}_h$ is decoded from $\hat{z}_h$ using the VAE decoder.

\noindent
\textbf{Regularization Loss.}
To preserve fidelity and perceptual consistency during adversarial refinement, we adopt a reconstruction loss that combines pixel-level MSE and perceptual LPIPS losses:
\begin{equation}
    \mathcal{L}_{\text{Rec}}
    =
    \mathcal{L}_{\text{MSE}}(\hat{x}_h,x_h)
    +
    \mathcal{L}_{\text{LPIPS}}(\hat{x}_h,x_h).
\label{eq:rec_loss}
\end{equation}

\noindent
\textbf{Total Loss.}
In the detail refinement stage, the student parameters $\theta$ are optimized with the following objective:
\begin{equation}
    \mathcal{L}_{\text{stu}}
    =
    \lambda_1 \mathcal{L}_{\text{ISC}}
    +
    \lambda_2 \mathcal{L}_{\text{Rec}}
    +
    \lambda_3 \mathcal{L}_{\text{VSD}}
    +
    \lambda_4 \mathcal{L}_{\text{adv}}^{\mathcal{G}},
\label{eq:total_loss}
\end{equation}
where $\lambda_1$, $\lambda_2$, $\lambda_3$, and $\lambda_4$ are balancing weights. The ISC term preserves the learned one-step noise-to-HR trajectory, while the reconstruction, VSD, and adversarial terms jointly improve fidelity, naturalness, and perceptual detail.
\section{Experimental Settings}
\label{sec:experimental settings}

% ----------------datasets-----------------
\noindent
\textbf{Training and Testing Datasets.}
We train SMFSR on LSDIR~\cite{lsdir} and the first 10K face images from FFHQ~\cite{ffhq}. Following common Real-ISR practice, we synthesize LR-HR training pairs using the degradation pipeline of Real-ESRGAN~\cite{2021Real}, and generate image captions with LLaVA~\cite{llava}. For evaluation, we use one synthetic benchmark, DIV2K-Val, and three real-world benchmarks, including RealSR~\cite{2020Toward}, DRealSR~\cite{drealsr}, and RealLQ250. RealLQ250 contains 250 real LR images with a resolution of $256\times256$ and has no paired HR references.

\begin{table*}
    \captionsetup{}
    \small
    \footnotesize
    \centering
    \renewcommand\arraystretch{1.25}
    \caption{Quantitative comparison against state-of-the-art methods across both synthetic and real-world datasets. The best and second best results of each metric are highlighted in {\color[HTML]{FF0000}\textbf{red}} and {\color[HTML]{0000FF}\textbf{blue}}, respectively.}
    
    % \vspace{-8pt}
    
    \setlength\tabcolsep{2.6pt}{
    \begin{tabular}{c|c|cc|ccccc|ccccccc}
    \toprule
    \multirow{3}{*}{\centering \textbf{Datasets}} & \multirow{3}{*}{\centering \textbf{Metrics}} & \multicolumn{13}{c}{\textbf{Methods}} \\ 
    \cmidrule(lr){3-16}
    & & \scriptsize \begin{tabular}[c]{@{}c@{}}\textbf{BSRGAN}\\ \end{tabular} & \scriptsize \begin{tabular}[c]{@{}c@{}}\textbf{SwinIR}\\ \end{tabular} & \scriptsize \begin{tabular}[c]{@{}c@{}}\textbf{StableSR}\\ \end{tabular} & \scriptsize \begin{tabular}[c]{@{}c@{}}\textbf{SUPIR}\\ \end{tabular} & \scriptsize \begin{tabular}[c]{@{}c@{}}\textbf{PASD}\\ \end{tabular} & \scriptsize \begin{tabular}[c]{@{}c@{}}\textbf{SeeSR}\\ \end{tabular} & \scriptsize \begin{tabular}[c]{@{}c@{}}\textbf{ResShift}\\ \end{tabular} & \scriptsize \begin{tabular}[c]{@{}c@{}}\textbf{S3Diff}\\ \end{tabular} & \scriptsize \begin{tabular}[c]{@{}c@{}}\textbf{OSEDiff}\\ \end{tabular} & \scriptsize \begin{tabular}[c]{@{}c@{}}\textbf{SinSR}\\ \end{tabular} & \scriptsize \begin{tabular}[c]{@{}c@{}}\textbf{CTMSR}\\ \end{tabular} & \scriptsize \begin{tabular}[c]{@{}c@{}}\textbf{InvSR}\\ \end{tabular} & \scriptsize \begin{tabular}[c]{@{}c@{}}\textbf{HYPIR} \\  \end{tabular} & \scriptsize \begin{tabular}[c]{@{}c@{}}\textbf{SMFSR} \\  \end{tabular} \\
    % & & \scriptsize StableSR\!~\cite{2024stablesr} & \scriptsize SUPIR~\cite{supir} & \scriptsize PASD~\cite{yang2023pasd} & \scriptsize SeeSR~\cite{wu2024seesr} & \scriptsize ResShift~\cite{resshift} & \scriptsize S3Diff~\cite{s3diff} & \scriptsize OSEDiff~\cite{2024OSEDiff} & \scriptsize SinSR~\cite{2024sinsr} & \scriptsize TSD-SR~\cite{dong2024tsd} & \scriptsize InvSR~\cite{invsr} & Ours \\
    \midrule
        \multirow{6}{*}{\footnotesize \textit{DIV2K}}
        & CLIPIQA $\uparrow$ & 0.5247 & 0.5338 & 0.6753 & 0.7046 & 0.6758 & 0.6865 & 0.5963 & 0.7001 & 0.6681 & 0.6488 & 0.6598 & \sbest{0.7181} & 0.6491 & \best{0.7545} \\ 
        & MUSIQ $\uparrow$ & 61.19 & 60.21 & 65.68 & 64.16 & 67.36 & \sbest{68.04} & 60.88 & 67.92 & 67.96 & 62.85 & 65.64 & 66.89 & 65.70 & \best{69.97} \\ 
        & MANIQA $\uparrow$ & 0.5068 & 0.5431 & 0.6141 & 0.5943 & 0.6132 & 0.6140 & 0.5338 & 0.5937  & 0.6131 & 0.5384 & 0.5166 & \sbest{0.6424} & 0.6119 & \best{0.6611} \\ 
        & PSNR $\uparrow$ & 24.58 & 23.93 & 22.61 & 22.43 & 23.15 & 23.01 & \sbest{24.71} & 23.53 & 23.72 & 24.41 & \best{24.87} & 22.90 & 22.25 & 22.18 \\ 
        & SSIM $\uparrow$ & \sbest{0.6269} & \best{0.6285} & 0.5712 & 0.5479 & 0.5512 & 0.6065 & 0.6183 & 0.5933 & 0.6109 & 0.6019 & 0.6262 & 0.5910 & 0.5721 & 0.5261 \\ 
        & LPIPS $\downarrow$ & 0.3351 & 0.3160 & 0.3113 & 0.3820 & 0.3543 & 0.3469 & 0.3402 & \sbest{0.2981} & \best{0.2941} & 0.3240 & 0.3028 & 0.3187 & 0.3041 & 0.3634 \\ 
        \midrule
        
        \multirow{6}{*}{\footnotesize \textit{DrealSR}}
       & CLIPIQA $\uparrow$ & 0.5093 & 0.4446 & 0.6282 & 0.6737 & 0.6812 & 0.6893 & 0.5404 & 0.7131 & 0.6958 & 0.6376 & 0.6517 & \sbest{0.7134} & 0.6392 & \best{0.7161} \\ 
        & MUSIQ $\uparrow$ & 57.15 & 52.73 & 58.58 & 58.66 & 63.23 & \sbest{64.75} & 52.37 & 63.93 & 64.69 & 55.38 & 59.76 & 63.99 & 61.07 & \best{65.93} \\ 
        & MANIQA $\uparrow$ & 0.4885 & 0.4750 & 0.5622 & 0.5517 & 0.5919 & 0.6014 & 0.4748 & 0.5719 & 0.5898 & 0.4907 & 0.4835 & 0.5920 & \sbest{0.6053} & \best{0.6220} \\ 
        & PSNR $\uparrow$ & \best{28.70} & 28.49 & 28.04 & 25.31 & 27.35 & 28.14 & \sbest{28.69} & 27.53 & 27.92 & 28.23 & 28.68 & 25.67 & 25.93 & 26.35 \\ 
        & SSIM $\uparrow$ & \sbest{0.8028} & \best{0.8044} & 0.7775 & 0.6558 & 0.7132 & 0.7712 & 0.7875 & 0.7491 & 0.7836 & 0.7468 & 0.7839 & 0.7131 & 0.7197 & 0.7021 \\ 
        & LPIPS $\downarrow$ & \sbest{0.2858} & \best{0.2743} & 0.2978 & 0.4122 & 0.3715 & 0.3142 & 0.3525 & 0.3109 & 0.2968 & 0.3707 & 0.3236 & 0.3537 & 0.3371 & 0.3860 \\ 
        \midrule 
        \multirow{6}{*}{\footnotesize \textit{RealSR}}
        & CLIPIQA $\uparrow$ & 0.5117 & 0.4365 & 0.6199 & 0.6619 & 0.6619 & 0.6673 & 0.5505 & 0.6731 & 0.6687 & 0.6224 & 0.6334 & \sbest{0.6789} & 0.6390 & \best{0.7065} \\ 
        & MUSIQ $\uparrow$ & 63.28 & 58.69 & 61.82 & 62.11 & 68.73 & \best{71.69} & 60.22 & 67.82 & 69.10 & 60.63 & 64.41 & 68.53 & 66.26 & \sbest{69.17} \\ 
        & MANIQA $\uparrow$ & 0.5419 & 0.5223 & 0.5702 & 0.5795 & \sbest{0.6467} & 0.6434 & 0.5402 & 0.6419 & 0.6326 & 0.5421 & 0.5268 & 0.6435 & 0.6436 & \best{0.6642} \\ 
        & PSNR $\uparrow$ & \sbest{26.37} & 26.30 & 24.85 & 23.70 & 25.12 & 25.21 & \best{26.38} & 25.18 & 25.15 & 25.64 & 25.99 & 24.13 & 22.83 & 23.14 \\ 
        & SSIM $\uparrow$ & \sbest{0.7651} & \best{0.7729} & 0.7043 & 0.6564 & 0.6889 & 0.7216 & 0.7347 & 0.7329 & 0.7341 & 0.7352 & 0.7546 & 0.7125 & 0.6783 & 0.6535 \\ 
        & LPIPS $\downarrow$ & \sbest{0.2656} & \best{0.2539} & 0.3029 & 0.3650 & 0.3391 & 0.3004 & 0.3159 & 0.2821 & 0.2921 & 0.3190 & 0.2896 & 0.2871 & 0.3087 & 0.3667 \\

        \midrule 
        \multirow{3}{*}{\footnotesize \textit{RealLQ250}}
        & CLIPIQA $\uparrow$ & 0.5689 & 0.5547 & 0.5156 & 0.5560 & 0.5574 & 0.6998 & 0.6132 & \sbest{0.7004} & 0.6723 & 0.6985 & 0.6700 & 0.6627 & 0.6899 & \best{0.7772} \\ 
        & MUSIQ $\uparrow$ & 63.51 & 63.37 & 57.48 & 63.14 & 62.04 & 65.15 & 59.49 & 69.19 & \sbest{69.55} & 63.80 & 68.01 & 65.82 & 69.05 & \best{72.21} \\ 
        & MANIQA $\uparrow$ & 0.5006 & 0.5335 & 0.5116 & 0.5762 & 0.5128 & 0.5807 & 0.5005 & 0.6016 & 0.5782 & 0.5152 & 0.5080 & 0.5819 & \sbest{0.6031} & \best{0.6410} \\ 
        \bottomrule 
    \end{tabular}}
    % \vspace{-4mm}
    \label{tab:scmsr_sota}
\end{table*}

\noindent
\textbf{Compared Methods.}
We compare SMFSR with three categories of Real-ISR methods: GAN-based methods, one-step diffusion-based methods, and multi-step diffusion-based methods. The GAN-based baselines include BSRGAN~\cite{2021Designing} and SwinIR~\cite{swinir}. The one-step diffusion-based methods include SinSR~\cite{2024sinsr}, OSEDiff~\cite{2024OSEDiff}, S3Diff~\cite{s3diff}, CTMSR~\cite{CTMSR}, HYPIR~\cite{HYPIR}, and InvSR~\cite{invsr}. The multi-step diffusion-based methods include StableSR~\cite{2024stablesr}, ResShift~\cite{resshift}, PASD~\cite{yang2023pasd}, SUPIR~\cite{supir}, and SeeSR~\cite{wu2024seesr}.

% ----------------evaluation metrics-----------------
\noindent
\textbf{Evaluation Metrics.}
We evaluate both reconstruction fidelity and perceptual quality. For paired benchmarks, PSNR and SSIM~\cite{ssim} are used to measure pixel-level fidelity, while LPIPS~\cite{lpips} measures perceptual similarity. For perceptual quality, especially on real-world unpaired benchmarks, we report three non-reference metrics: MUSIQ~\cite{2021MUSIQ}, MANIQA~\cite{2022MANIQA}, and CLIPIQA~\cite{clipiqa}.

% ----------------implementation details-----------------
\noindent
\textbf{Implementation Details.}
We adopt a two-stage training strategy. First, we train a teacher model with standard flow matching using AdamW~\cite{2019decoupledweight} and a fixed learning rate of $5\times10^{-5}$. The teacher is trained for 80K iterations on 4 NVIDIA 5880 GPUs with a batch size of 32. The student model is then initialized from the pre-trained teacher and optimized with $\mathcal{L}_{\text{ISC}}$ for 40K iterations using a fixed learning rate of $5\times10^{-5}$ and a batch size of 16. This stage learns an LR-conditioned one-step average-velocity mapping from random Gaussian noise to HR latents. In the second stage, we further optimize the student for 10K iterations with a batch size of 8 using the GAN-based refinement objectives. The loss weights $\lambda_1$, $\lambda_2$, $\lambda_3$, and $\lambda_4$ are set to 1.0, 1.0, 1.0, and 0.5, respectively. The probability of using the boundary branch $p$ is set to 0.6. The teacher is built upon Stable Diffusion 3.5 and initialized from SD3.5-Medium~\cite{sd3}. The student follows the same architecture as the teacher, except that its input and timestep formulation are modified to support two timesteps.

\subsection{Comparison with State-of-the-Arts}
\label{comparison}

% ----------------quantitative comparisons-----------------
\begin{figure*}[t!]
    \centering
    \includegraphics[width=\linewidth]{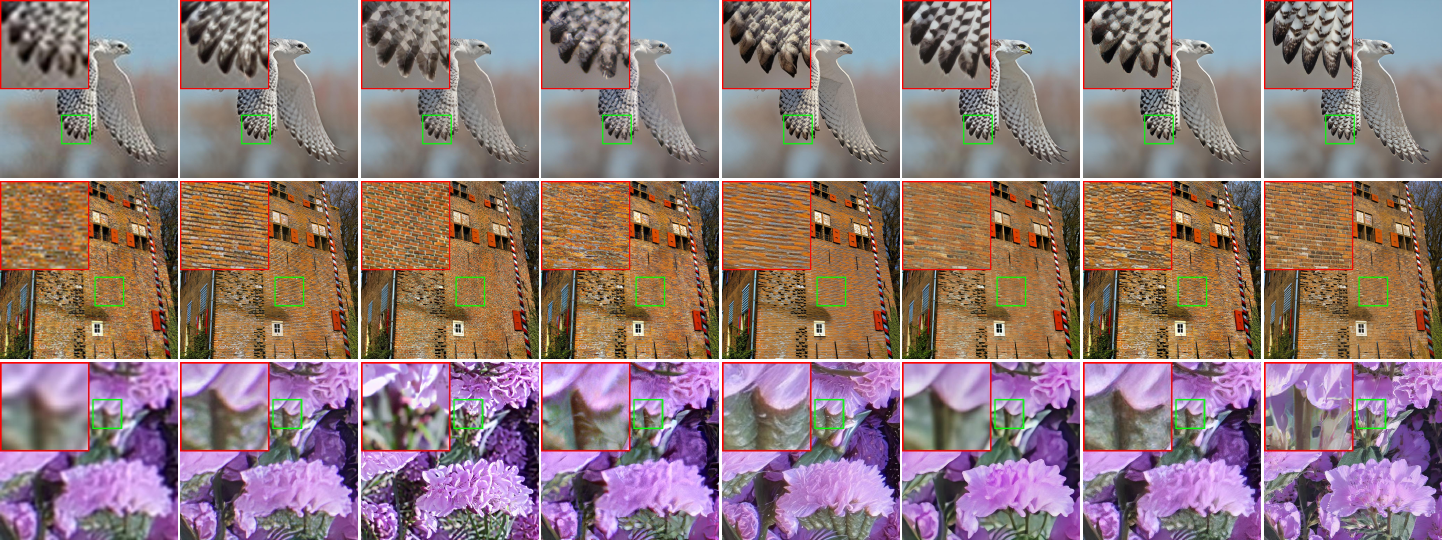}
    % \vspace{3pt}
    \scriptsize
    \makebox[0.125\linewidth][c]{(a) LQ}%
    \makebox[0.125\linewidth][c]{(b) StableSR-s200}%
    \makebox[0.125\linewidth][c]{(c) SUPIR-s50}%
    \makebox[0.125\linewidth][c]{(d) ResShift-s15}%
    \makebox[0.125\linewidth][c]{(e) InvSR-s1}%
    \makebox[0.125\linewidth][c]{(f) OSEDiff-s1}%
    \makebox[0.125\linewidth][c]{(g) CTMSR-s1}%
    \makebox[0.125\linewidth][c]{(h) Ours-s1}%
    % \vspace{-2mm}
    \caption{
        Visual comparison of different diffusion-based Real-ISR methods, where ``s'' denotes the number of inference steps.
    }
    \label{fig:visual comparisons}
    \vspace{-4mm}
\end{figure*}

\noindent
\textbf{Quantitative Comparisons.}
Tab.~\ref{tab:scmsr_sota} reports quantitative comparisons with state-of-the-art Real-ISR methods on four benchmarks. SMFSR achieves the best overall performance in non-reference perceptual quality, obtaining the highest MUSIQ, MANIQA, and CLIPIQA scores on both synthetic and real-world datasets. The advantage is particularly clear on the challenging RealLQ250 benchmark, where no paired HR references are available and perceptual realism becomes the primary evaluation criterion. These results demonstrate that the proposed LR-conditioned SplitMeanFlow and GAN refinement effectively improve realistic texture synthesis under one-step inference.

SMFSR is less competitive on full-reference metrics such as PSNR and SSIM. This is expected, as full-reference metrics favor pixel-wise alignment with a specific reference image, whereas Real-ISR is inherently a one-to-many problem. Methods that synthesize more realistic high-frequency details may deviate from the reference at the pixel level, leading to lower fidelity scores despite better perceptual quality. This observation is consistent with prior Real-ISR studies~\cite{supir,wu2024seesr} and the well-known perception-distortion trade-off~\cite{perception,Blau_2018_CVPR}.

\noindent
\textbf{Qualitative Comparisons.}
Fig.~\ref{fig:visual comparisons} presents visual comparisons among different methods. SMFSR produces sharper structures and more natural details, while competing methods often suffer from over-smoothing, blurred textures, or structural artifacts. The advantage is especially evident in regions with fine patterns and complex structures, where our method better preserves local detail clarity and global structural coherence. This improvement comes from two key designs: LR-conditioned SplitMeanFlow preserves the noise-started generative formulation for stochastic detail synthesis, and the GAN-based refinement stage further enhances high-frequency textures and perceptual realism. In contrast, existing one-step methods that directly restore HR images from LR inputs often struggle to recover subtle visual details.

% ----------------complexity comparisons-----------------
\noindent
\textbf{Complexity Comparisons.}
Tab.~\ref{tab:complexity_comparison} compares the complexity of representative Real-ISR methods in terms of inference steps, runtime, and trainable parameters. All methods are evaluated on the $\times 4$ SR task with $128\times128$ LQ inputs using a single NVIDIA 5880 GPU. StableSR and SeeSR rely on SD2~\cite{sd21-base} and require multi-step denoising, resulting in high inference cost. SUPIR is built upon SDXL~\cite{2023sdxl} and contains the largest number of trainable parameters. S3Diff and InvSR adopt SD-Turbo~\cite{sd-tubro} and have comparable parameter sizes. OSEDiff is based on SD2.1-Base~\cite{sd21-base} and achieves the fastest runtime of 0.19 seconds.

SMFSR uses SD3.5-Medium~\cite{sd3} as the generative prior and therefore contains more trainable parameters than several one-step baselines. Nevertheless, it achieves a fast inference time of 0.22 seconds, which is only slightly slower than OSEDiff. This shows that SMFSR preserves the noise-started generation paradigm of diffusion models without sacrificing the efficiency advantage of one-step inference. The reported runtime excludes the overhead of text extraction. As shown in Tab.~\ref{tab:caption_type}, the choice of caption extractor has little influence on performance.

% \begin{table}[tbp!]
\begin{table}
\centering
\scriptsize
 \renewcommand\arraystretch{1.2}
\caption{Complexity comparison of different methods. All evaluations are measured on an NVIDIA RTX 5880 GPU, where each method generates $512 \times 512$ results from $128 \times 128$ inputs.}
\vspace{-1mm}
\setlength\tabcolsep{4.5pt}
\begin{tabular}{l|lccc}
\toprule
\textbf{Methods} & \begin{tabular}[c]{@{}c@{}}\textbf{Base Model}\end{tabular} & \begin{tabular}[c]{@{}c@{}}\textbf{Sample}\\ \textbf{Steps}\end{tabular} &  \begin{tabular}[c]{@{}c@{}}\textbf{Inference}\\ \textbf{Time (s)}\end{tabular} & \begin{tabular}[c]{@{}c@{}}\textbf{Trainable}\\ \textbf{Param (M)}\end{tabular} \\
\midrule
StableSR & SD2 & 200 & 14.15 & 153.3  \\
SUPIR & SDXL & 50 & 16.87 & 1,331.0 \\
SeeSR & SD2 & 50 & 6.36 & 751.7 \\
ResShift & Diffusion & 15 & 1.47 & 118.0 \\
\midrule
S3Diff & SD-Turbo & 1 & 0.76 & 34.5 \\
% SinSR & & 1 & 0.12 & 118.0 \\
InvSR & SD-Turbo & 1 & 0.27 & 33.8  \\
OSEDiff & SD2.1-Base & 1 & 0.19 & 8.5  \\
CTMSR & Diffusion & 1 & 1.31 & 171.5 \\
\textbf{SMFSR} & SD3.5-M & 1 & 0.22 & 366.2 \\
% SCMSR (w/ LLaVA) & 1 & 1.2 & 366.22 \\
\bottomrule
\end{tabular}

\label{tab:complexity_comparison}
\vspace{-4mm}
\end{table}

% % \begin{table}[tbp!]
% \begin{table}[]
% \centering
% \caption{Complexity comparison of different methods. All methods are evaluated on $512 \times 512$ resolution. The inference time is measured on an 5880 GPU.}
% \setlength\tabcolsep{2pt}
% \begin{tabular}{l|ccc}
% \toprule
% \textbf{Method $|$ Prompt Model} & \begin{tabular}[c]{@{}c@{}}\textbf{Inference}\\ \textbf{Steps}\end{tabular} &  \begin{tabular}[c]{@{}c@{}}\textbf{Inference}\\ \textbf{Times (s)}\end{tabular} & \begin{tabular}[c]{@{}c@{}}\textbf{Trainable}\\ \textbf{Param (M)}\end{tabular} \\
% \midrule
% StableSR~\cite{2024stablesr} (None) & 200 & 14.15 & 150.0 \\
% SUPIR~\cite{supir} (LLaVA) & 50 & 16.87 & 1,331 \\
% SeeSR~\cite{wu2024seesr} (DAPE) & 50 & 6.36 & 749 \\
% ResShift~\cite{resshift} (None)& 15 & 1.47 & 118 \\
% \midrule
% S3Diff~\cite{s3diff} (Fix) & 1 & 0.76 & 34.46 \\
% SinSR~\cite{2024sinsr} (None) & 1 & 0.12 & 118 \\
% InvSR~\cite{invsr} (Fix) & 1 & 0.27 & 33.8 \\
% OSEDiff~\cite{2024OSEDiff} (DAPE) & 1 & 0.19 & 8.5 \\
% CTMSR~\cite{CTMSR} (No Prompt) & 1 & 1.31 & 171.52 \\
% SCMSR (offline LLaVA) & 1 & 0.22 & 366.22 \\
% % SCMSR (w/ LLaVA) & 1 & 1.2 & 366.22 \\
% \bottomrule
% \end{tabular}

% \label{tab:complexity_comparison}
% \vspace{-0.3cm}
% \end{table}

% ----------------Ablation Study-----------------
\begin{figure*}
    \centering
    \includegraphics[width=0.98\linewidth]{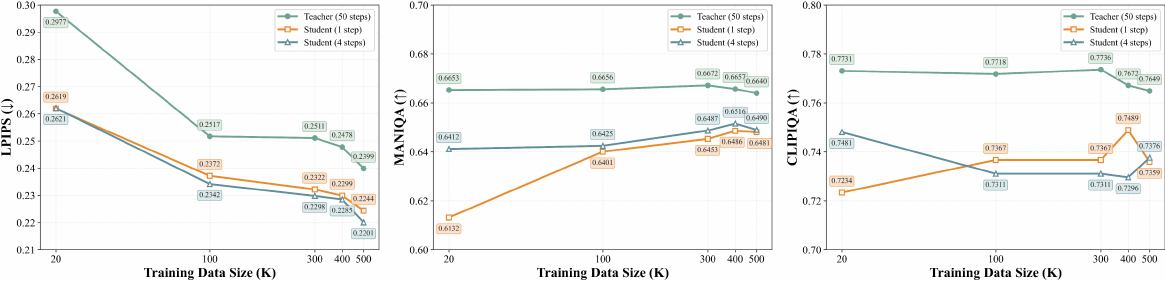}
    % \vspace{-9pt}
    \caption{
        Teacher vs. student performance under different training set sizes. As the amount of training data increases, the distilled student surpasses the teacher in LPIPS, but remains consistently behind in CLIPIQA and MANIQA, indicating limited perceptual detail refinement after one-step trajectory approximation.
    }
    \label{fig:teacher_vs_student}
    % \vspace{-4mm}
\end{figure*}
\subsection{ISC Exhibits Limited Detail Refinement}
\label{sec:limited_detail_recovery_main}
To analyze the capacity and limitation of Interval Splitting Consistency (ISC) for one-step generation, we first consider a simplified degradation setting that includes only random resizing and JPEG compression. Specifically, LR inputs are synthesized from HR images by random resizing followed by JPEG compression. For resizing, we randomly choose up-sampling, down-sampling, or keeping the original resolution with probabilities of 0.2, 0.7, and 0.1, respectively. The resizing scale factor is uniformly sampled from $[0.5,1.5]$, and the interpolation method is randomly selected from area, bilinear, and bicubic. JPEG compression is then applied with the quality factor uniformly sampled from $[30,50]$. Experiments are conducted on LSDIR, and we construct a 100-image test set using the same degradation process as training.

Fig.~\ref{fig:teacher_vs_student} shows two important observations. First, as the training set size increases, the one-step student gradually saturates. Increasing the number of inference steps from one to four brings only marginal gains, indicating that ISC can effectively learn a compact average-velocity trajectory for one-step generation. Second, even after saturation, a clear gap remains between the student and the multi-step teacher in non-reference perceptual metrics, including CLIPIQA and MANIQA. This suggests that the remaining limitation is not simply caused by insufficient inference steps, but rather by the intrinsic difficulty of representing progressive detail formation with a single average-velocity prediction.

This finding directly motivates the GAN-based refinement stage of SMFSR. From this perspective, our method can be interpreted as a noise-to-structure-to-detail framework: LR-conditioned SplitMeanFlow first recovers the main content through noise-started one-step generation, while GAN refinement further enhances perceptual details and texture realism.

\begin{figure*}
    \centering
    \includegraphics[width=\linewidth]{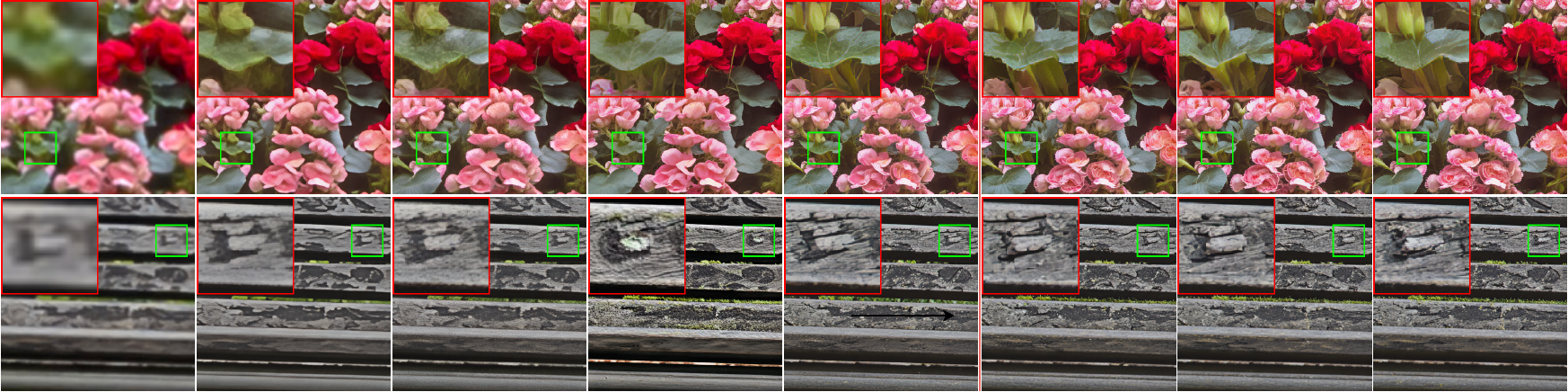}
    % \vspace{3pt}
    \scriptsize
    \makebox[0.125\linewidth][c]{\text{(a) Zoomed LQ}}%
    \makebox[0.125\linewidth][c]{\text{(b) Rec.}}%
    \makebox[0.125\linewidth][c]{\text{(c) Rec.+VSD}}%
    \makebox[0.125\linewidth][c]{\text{(d) Rec.+GAN}}%
    \makebox[0.125\linewidth][c]{(e) \textbf{Full (Seed 1)}}%
    \makebox[0.125\linewidth][c]{(f) \textbf{Full (Seed 2)}}%
    \makebox[0.125\linewidth][c]{(g) \textbf{Full (Seed 3)}}%
    \makebox[0.125\linewidth][c]{(h) \textbf{Full (Seed 4)}}%
    % \vspace{-1mm}
    \caption{
        Visual comparison of different losses in the GAN-based refinement stage. The last four columns show results generated with different random seeds, demonstrating the stochasticity preserved by noise-started one-step generation.
    }
    \label{fig:loss_visual_comparisons}
    % \vspace{-3mm}
\end{figure*}

% ----------------Model Stability and Diversity-----------------
\subsection{Model Stability and Diversity}
\label{subsec:stability_diversity}

Since SMFSR preserves the noise-started generation paradigm, it naturally supports diverse outputs for the same LR input. As shown in the last four columns of Fig.~\ref{fig:loss_visual_comparisons}, different random seeds produce subtle variations in local textures while maintaining consistent global structures. This behavior is consistent with diffusion-based generation, where different noise realizations can yield diverse yet plausible HR results under the same LR condition. Such stochasticity is difficult to obtain for one-step methods that directly map LR inputs to HR outputs.

We further evaluate model stability by fixing all hyperparameters and varying only the random seed. Specifically, we randomly sample 20 seeds and report the mean and standard deviation of the metrics in Tab.~\ref{tab:mean_std}. We also measure LPIPS across different seeds by using the result generated with seed 1 as the reference, as reported in Tab.~\ref{tab:seed_lpips}. The results show that SMFSR maintains low metric variance while producing reasonable local variations, demonstrating that the proposed noise-started one-step generation is both stable and diverse.

\begin{table*}
\centering
\scriptsize
\renewcommand\arraystretch{1.0}
\caption{Mean and standard deviation of quantitative metrics over 20 random seeds.}
\label{tab:mean_std}
\resizebox{0.95\linewidth}{!}{
\begin{tabular}{l|cccccc}
\toprule
\textbf{Datasets} & \textbf{PSNR$\uparrow$} & \textbf{SSIM$\uparrow$} & \textbf{LPIPS$\downarrow$} & \textbf{MUSIQ$\uparrow$} & \textbf{MANIQA$\uparrow$} & \textbf{CLIPIQA$\uparrow$} \\
\midrule
\textbf{RealSR} &
23.2420$\pm$0.0095 &
0.6571$\pm$0.0006 &
0.3574$\pm$0.0074 &
68.9506$\pm$0.1750 &
0.6630$\pm$0.0008 &
0.7087$\pm$0.0027 \\
\textbf{DrealSR} &
26.3852$\pm$0.0198 &
0.7033$\pm$0.0006 &
0.3856$\pm$0.0009 &
66.1328$\pm$0.1470 &
0.6215$\pm$0.0010 &
0.7161$\pm$0.0031  \\
\bottomrule
\end{tabular}
}
\end{table*}

\begin{table}
\centering
\scriptsize
\renewcommand\arraystretch{1.2}
\caption{LPIPS metrics evaluated across different random seeds, taking the results from seed 1 as the reference.}
\label{tab:seed_lpips}
% \resizebox{\linewidth}{!}{
\begin{tabular}{l|cccccc}
\toprule
\textbf{Datasets} & \textbf{Seed 2} & \textbf{Seed 3} & \textbf{Seed 4} & \textbf{Seed 5} & \textbf{Seed 6} & \textbf{Seed 7} \\
\midrule
\textbf{RealSR} &
0.1227 &
0.1219 &
0.1236 &
0.1221 &
0.1224 &
0.1219  \\
\textbf{DrealSR} &
0.1353 &
0.1380 &
0.1402 &
0.1378 &
0.1376 &
0.1367 \\
\bottomrule
\end{tabular}
% }
\end{table}

% ----------------Ablation Study-----------------

\subsection{Ablation Study}
\label{sec:ablation_study}

\begin{table}
\centering
\scriptsize
\caption{Comparison of different losses on the RealSR benchmark. Rec. denotes the reconstruction loss using $\mathcal{L}_\text{LPIPS}$ and $\mathcal{L}_\text{MSE}$.
}
% \vspace{-2mm}
% \renewcommand{\arraystretch}{1.5}
\resizebox{\linewidth}{!}{
\begin{tabular}{l|cccc}
\toprule
\textbf{Method} & \textbf{LPIPS $\downarrow$} & \textbf{MUSIQ $\uparrow$} & \textbf{MANIQA $\uparrow$} & \textbf{CLIPIQA $\uparrow$} \\ \midrule 
\textit{Rec.} & \best{0.2664} & 66.16 & 0.5981 & 0.5692 \\
\textit{Rec. + VSD} & 0.3049 & 67.92 & 0.6265 & 0.6591 \\
\textit{Rec. + GAN} & 0.3482 & 68.16 & 0.6538 & 0.6916 \\
\textit{Full} & 0.3667 & \best{69.17} & \best{0.6642} & \best{0.7065} \\

\bottomrule
\end{tabular}%
}
\label{tab:ablation_loss}
\vspace{-4mm}
\end{table}

% ------------- loss ablation ---------------
\noindent
\textbf{Loss Ablation Study.}
We ablate the contribution of each loss term in the GAN-based refinement stage in Tab.~\ref{tab:ablation_loss}. Using only the reconstruction loss improves full-reference metrics but substantially degrades non-reference perceptual metrics, indicating that pixel-level supervision alone is insufficient to compensate for the loss of progressive detail refinement in one-step generation. Adding either VSD or GAN loss significantly improves non-reference metrics. In particular, the GAN loss brings larger gains on CLIPIQA, confirming the importance of adversarial supervision for high-frequency detail synthesis. Combining all loss terms achieves the best overall perceptual performance. The qualitative results in Fig.~\ref{fig:loss_visual_comparisons} further show that GAN loss restores sharper local details than VSD alone, while their combination produces more realistic and visually pleasing results.

\noindent
\textbf{CFG Scale for Boundary Consistency Loss.}
We study the effect of the teacher CFG scale $w$ in \textit{stage 1} when applying boundary consistency. As shown in Tab.~\ref{tab:cfg_scale}, increasing the CFG scale slightly improves perceptual metrics such as MUSIQ and CLIPIQA, but noticeably degrades fidelity. In contrast, disabling CFG yields the best fidelity while maintaining competitive perceptual performance. Therefore, we do not use CFG for the boundary consistency loss in \textit{stage 1}, which provides a stable initialization for the subsequent GAN-based perceptual optimization in \textit{stage 2}.

\noindent
\textbf{Multi-step Inference is Redundant After \textit{Stage 1}.}
We further examine whether additional inference steps remain beneficial after \textit{stage 1}. Specifically, we increase the number of steps from 1 to 2 and 4 while keeping all other settings unchanged. As shown in Tab.~\ref{tab:ablation_after_stage1}, the student already saturates at a single step, and additional steps do not improve reconstruction quality. This indicates that ISC training has learned an effective noise-to-HR trajectory through the average-velocity formulation. Meanwhile, compared with the multi-step teacher, the student still exhibits weaker perceptual details. This confirms that the main limitation after \textit{stage 1} is not the number of inference steps, but the difficulty of representing progressive texture formation with a single average-velocity prediction. This limitation is therefore addressed by the GAN-based refinement stage.

\begin{table}
\centering
\scriptsize
\caption{Impact of teacher CFG scale on performance with boundary consistency loss in \textit{stage 1} on the RealSR benchmark.
}
% \vspace{-2mm}
% \renewcommand{\arraystretch}{1.5}
% \resizebox{\linewidth}{!}{
\begin{tabular}{l|cccc}
\toprule
\textbf{CFG Scale} & \textbf{LPIPS $\downarrow$} & \textbf{MUSIQ $\uparrow$} & \textbf{MANIQA $\uparrow$} & \textbf{CLIPIQA $\uparrow$} \\ \midrule 
\textit{w/o CFG} &\best{0.3083} & 65.56 & 0.5708 & 0.6367 \\
\textit{4.5} & 0.3661 & 66.04 & \best{0.6067} & 0.6383 \\
\textit{7.5} & 0.3823 & \best{67.07} & 0.5917 & \best{0.6446} \\

\bottomrule
\end{tabular}%
% }
\label{tab:cfg_scale}
\vspace{-2mm}
\end{table}
% \begin{table}
% \centering
% \scriptsize
% \caption{Ablation on inference steps after \textit{stage 1}. Performance saturates at a single step, with no consistent gains from increasing the number of steps.}
% \vspace{-1mm}
% % \renewcommand{\arraystretch}{1.5}
% % \resizebox{\linewidth}{!}{
% \begin{tabular}{c|c|cccc}
% \toprule
% \textbf{Model} & \textbf{Steps} & \textbf{LPIPS $\downarrow$} & \textbf{MUSIQ $\uparrow$} & \textbf{MANIQA $\uparrow$} & \textbf{CLIPIQA $\uparrow$} \\ \midrule 
% \textbf{Teacher} & 50 & 0.3083 & 65.56 & 0.5708 & \best{0.6367} \\
% \textbf{Student} & 1 & 0.3083 & 65.56 & 0.5708 & \best{0.6367} \\
% \textbf{Student} & 2 & \best{0.3064} & \best{65.59} & 0.5959 & 0.6323 \\
% \textbf{Student} & 4 & 0.3087 & 65.50 & \best{0.6034} & 0.6224 \\

% \bottomrule
% \end{tabular}%
% % }
% \label{tab:ablation_after_stage1}
% \vspace{-3mm}
% \end{table}

\begin{table}
\centering
\scriptsize
\caption{Ablation on the student model after \textit{stage 1}, together with a comparison to the multi-step teacher. The student performance saturates at a single step, with no consistent gains from increasing the number of inference steps.}
\vspace{-1mm}
\begin{tabular}{c|c|cccc}
\toprule
\textbf{Model} & \textbf{Steps} & \textbf{LPIPS $\downarrow$} & \textbf{MUSIQ $\uparrow$} & \textbf{MANIQA $\uparrow$} & \textbf{CLIPIQA $\uparrow$} \\ \midrule
\textbf{Teacher} & 50 & 0.3991 & \best{69.12} & \best{0.6608} & \best{0.6859} \\
\midrule
\multirow{3}{*}{\textbf{SMFSR}} 
& 1 & 0.3083 & 65.56 & 0.5708 & \sbest{0.6367} \\
& 2 & \sbest{0.3064} & \sbest{65.59} & 0.5959 & 0.6323 \\
& 4 & 0.3087 & 65.50 & \sbest{0.6034} & 0.6224 \\
\bottomrule
\end{tabular}
\label{tab:ablation_after_stage1}
% \vspace{-3mm}
\end{table}

\begin{table}[]
% \begin{wraptable}{l}{0.45\textwidth}
% \setlength{\abovecaptionskip}{0.1cm}
\setlength{\tabcolsep}{2pt}
\centering
\scriptsize
% \vspace{-2mm}
\caption{Comparison of different text prompt extractors.}
% \resizebox{0.48\textwidth}{!}{%
\begin{tabular}{c|c|ccccc}
\toprule
\textbf{Datasets} & \textbf{Prompt Model} & \textbf{PSNR $\uparrow$}& \textbf{SSIM $\uparrow$}& \textbf{LPIPS $\downarrow$} & \textbf{MUSIQ $\uparrow$} & \textbf{MANIQA $\uparrow$} \\ \midrule
\multirow{3}{*}{\textbf{RealSR}}
& NULL & 23.25 & 0.6582 & 0.3641 & 68.78 & 0.6609 \\
& LLaVA-v1.5 & 23.14 & 0.6535 & 0.3667 & \best{69.17} & \best{0.6642} \\
& DAPE & \best{23.33} & \best{0.6612} & \best{0.3564} & 69.09 & 0.6620 \\
\midrule
\multirow{3}{*}{\textbf{DrealSR}}
& NULL & 26.38 & 0.7063 & 0.3866 & 65.44 & 0.6190 \\
& LLaVA-v1.5 & 26.35 & 0.7021 & 0.3860 & \best{65.93} & \best{0.6220} \\
& DAPE & \best{26.63} & \best{0.7174} & \best{0.3736} & 65.65 & 0.6174 \\
\bottomrule
\end{tabular}%
% }
\label{tab:caption_type}
\vspace{-4mm}
\end{table}
% \end{wraptable}

\noindent
\textbf{Comparison on Text Prompt Extractors.}
We evaluate the effect of different text prompt extractors on DRealSR and RealSR. Specifically, we consider three prompt settings: (1) no text prompt (NULL), (2) degradation-aware tag-style prompts generated by DAPE from SeeSR~\cite{wu2024seesr}, and (3) long caption-style prompts generated by LLaVA-v1.5~\cite{llava}. As shown in Tab.~\ref{tab:caption_type}, SMFSR achieves comparable performance across different prompt types, indicating that it is not sensitive to the choice of text prompt extractor. We attribute this robustness to the text-dropout strategy during training, where the text prompt is randomly dropped with a probability of 0.2. This encourages the model to rely primarily on the LR image condition rather than overfitting to potentially noisy or inaccurate textual descriptions.
\section{Conclusion and Discussion}
\label{sec:conclusion}
This paper presented SMFSR, a noise-started one-step framework for real-world image super-resolution. In contrast to existing one-step methods that directly mapped LR inputs to HR outputs, SMFSR preserved the random-noise starting point of diffusion models and learned an LR-conditioned noise-to-HR mapping through SplitMeanFlow. This formulation retained stochastic generation, enabling diverse yet plausible HR outputs for the same LR input, while still allowing efficient single-step inference. Interval Splitting Consistency distilled the multi-step generative trajectory into a single average-velocity prediction, and a GAN-based refinement stage further compensated for the limited progressive detail refinement through adversarial supervision, VSD, and reconstruction regularization. Extensive experiments demonstrated that SMFSR achieved state-of-the-art perceptual quality among one-step diffusion-based Real-ISR methods while retaining fast inference.
The preserved stochasticity suggested a promising direction for future preference-based optimization in one-step Real-ISR, such as DPO~\cite{DPO} and DiffusionNFT~\cite{diffusionnft}.
{
    \small
    % \FloatBarrier
    \bibliographystyle{ieeenat_fullname}
    \bibliography{main}
}

\end{document}